\documentclass[10pt,twocolumn,letterpaper]{article}

\usepackage[pagenumbers]{cvpr} % To force page numbers, e.g. for an arXiv version
\usepackage[accsupp]{axessibility}  % Improves PDF readability for those with disabilities.

\usepackage{graphicx}
\usepackage{amsmath}
\usepackage{amssymb}
\usepackage{booktabs}
\usepackage{amsfonts}

\usepackage{bbding}
\usepackage{multirow}
\usepackage{color}

\usepackage[pagebackref,breaklinks,colorlinks]{hyperref}

\usepackage[capitalize]{cleveref}
\crefname{section}{Sec.}{Secs.}
\Crefname{section}{Section}{Sections}
\Crefname{table}{Table}{Tables}
\crefname{table}{Tab.}{Tabs.}

\def\cvprPaperID{1455} % *** Enter the CVPR Paper ID here
\def\confName{CVPR}
\def\confYear{2023}

\begin{document}

%\title{Diverse Embedding Expansion and A New Benchmark for \\
%Visible-Infrared Person Re-Identification}
\title{Diverse Embedding Expansion Network and Low-Light Cross-Modality \\
Benchmark for Visible-Infrared Person Re-identification}

\author{Yukang Zhang\textsuperscript{\rm 1,2}, Hanzi Wang\textsuperscript{\rm 1,2,3}\thanks{Corresponding author.}\\
\textsuperscript{\rm 1}Fujian Key Laboratory of Sensing and Computing for Smart City, \\School of Informatics, Xiamen University, 361005, P.R. China.\\
\textsuperscript{\rm 2}Key Laboratory of Multimedia Trusted Perception and Efficient Computing, \\Ministry of Education of China, Xiamen University, 361005, P.R. China.\\
\textsuperscript{\rm 3}Shanghai Artificial Intelligence Laboratory, Shanghai, 200232, China.\\
{\tt\small zhangyk@stu.xmu.edu.cn, hanzi.wang@xmu.edu.cn}
%\and
%{\tt\small secondauthor@i2.org
%}
\and
}

\maketitle
%%%%%%%%% ABSTRACT
\begin{abstract}
For the visible-infrared person re-identification (VIReID) task, one of the major challenges is the modality gaps between visible (VIS) and infrared (IR) images. However, the training samples are usually limited, while the modality gaps are too large, which leads that the existing methods cannot effectively mine diverse cross-modality clues. To handle this limitation, we propose a novel augmentation network in the embedding space, called diverse embedding expansion network (DEEN). The proposed DEEN can effectively generate diverse embeddings to learn the informative feature representations and reduce the modality discrepancy between the VIS and IR images. Moreover, the VIReID model may be seriously affected by drastic illumination changes, while all the existing VIReID datasets are captured under sufficient illumination without significant light changes. Thus, we provide a low-light cross-modality (LLCM) dataset, which  contains 46,767 bounding boxes of 1,064 identities captured by 9 RGB/IR cameras. Extensive experiments on the SYSU-MM01, RegDB and LLCM datasets show the superiority of the proposed DEEN over several other state-of-the-art methods. The code and dataset are released at: \textcolor{magenta}{\url{https://github.com/ZYK100/LLCM}}
\end{abstract}

%%%%%%%%% BODY TEXT

\section{Introduction}
\label{sec:intro}

\begin{figure}[t]
\includegraphics[height=4.4cm,width=8.2cm]{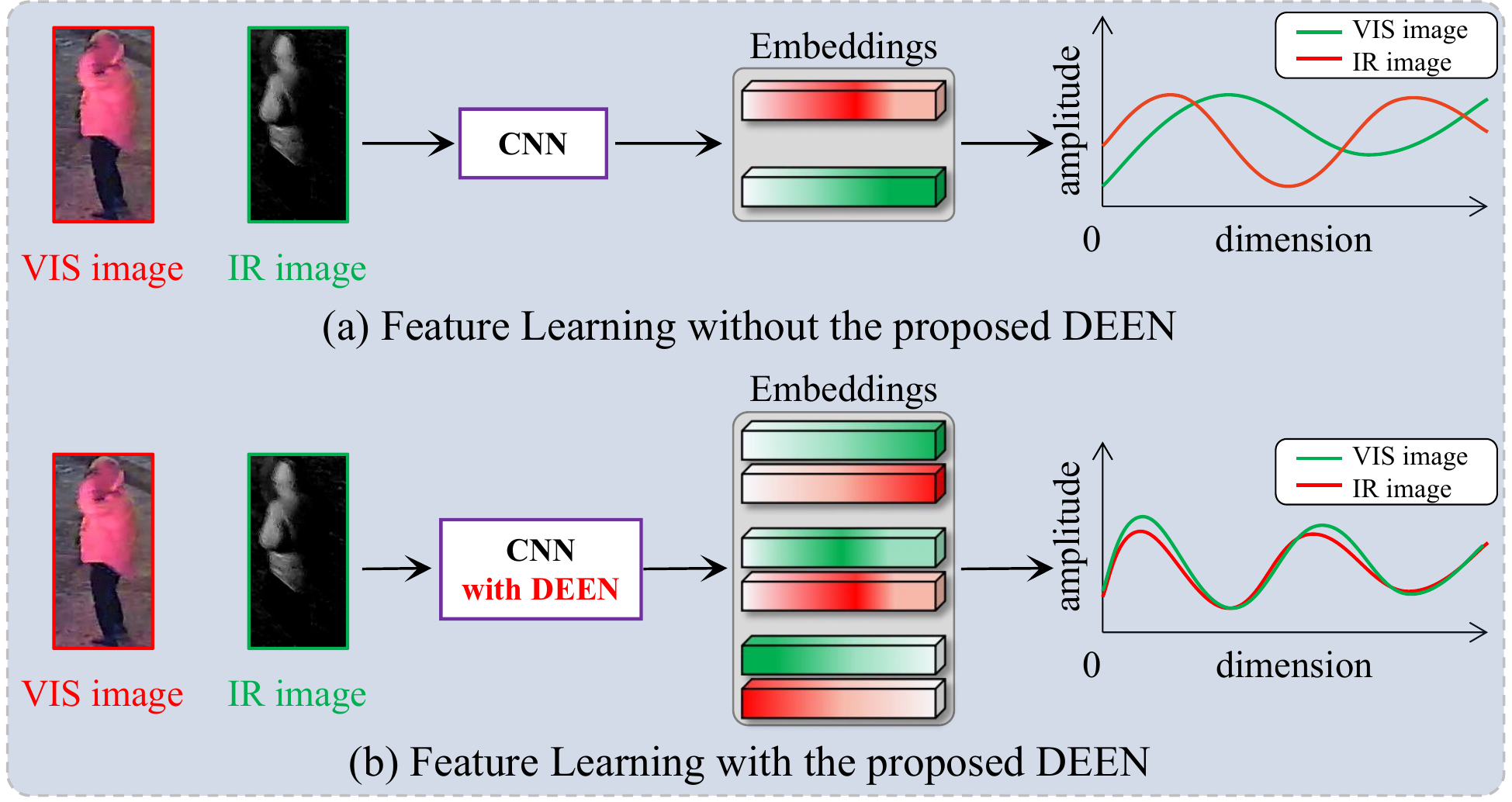}
\vspace{-0.2cm}
\caption{Motivation of the proposed DEEN, which aims to generate diverse embeddings to make the network focus on learning with the informative feature representations to reduce the modality gaps between the VIS and IR images.}
\vspace{-0.3cm}
\label{img:img1}
\end{figure}
Person re-identification (ReID) aims to match a given person with gallery images captured by different cameras \cite{Cho_2022_CVPR, Zheng_2021_ICCV, Gu_2022_CVPR}. Most existing ReID methods \cite{Zhang_2021_CVPR, Pu_2021_CVPR, Yan_2021_ICCV, Wang_2022_CVPR, Tan_2022_ACM} only focus on matching RGB images captured by visible cameras at daytime. %Thanks to the rapid development of deep learning, various deep end-to-end methods have been studied, which greatly enhances the performance of ReID model. 
However, these methods may fail to achieve encouraging results when visible cameras cannot effectively capture person's information under complex conditions, such as at night or low-light environments. To solve this problem, some visible (VIS)-infrared (IR) person re-identification (VIReID) methods \cite{LiuJialun_2022_CVPR, Yang_2022_CVPR, Zhang_2022_CVPR, Yang_2022_ACM} have been proposed to retrieve the VIS (IR) images according to the corresponding IR (VIS) images. 

Compared with the widely studied person ReID task, the VIReID task is much more challenging due to the additional cross-modality discrepancy between the VIS and IR images \cite{ye2020deep, zhang2021towards, wei2020co, zhao2021joint}. 
Typically, there are two popular types of methods to reduce this modality discrepancy. One type is the feature-level methods \cite{yang2020mining, ye2018hierarchical, dai2018cross, hao2019hsme, wu2020rgb, lu2020cross}, which try to project the VIS and IR features into a common embedding space, where the modality discrepancy can be minimized. However, the large modality discrepancy makes these methods difficult to project the cross-modality images into a common feature space directly. The other type is the image-level methods \cite{wang2019learning, wang2020cross, wang2019aligngan, choi2020hi}, which aim to reduce the modality discrepancy by translating an IR (or VIS) image into its VIS (or IR) counterpart by using the GANs \cite{goodfellow2014generative}. Despite their success in reducing the modality gaps, the generated cross-modality images are usually accompanied by some noises due to the lack of the VIS-IR image pairs.

\begin{figure}[t]
\centerline{\includegraphics[height=5.7cm,width=8.4cm]{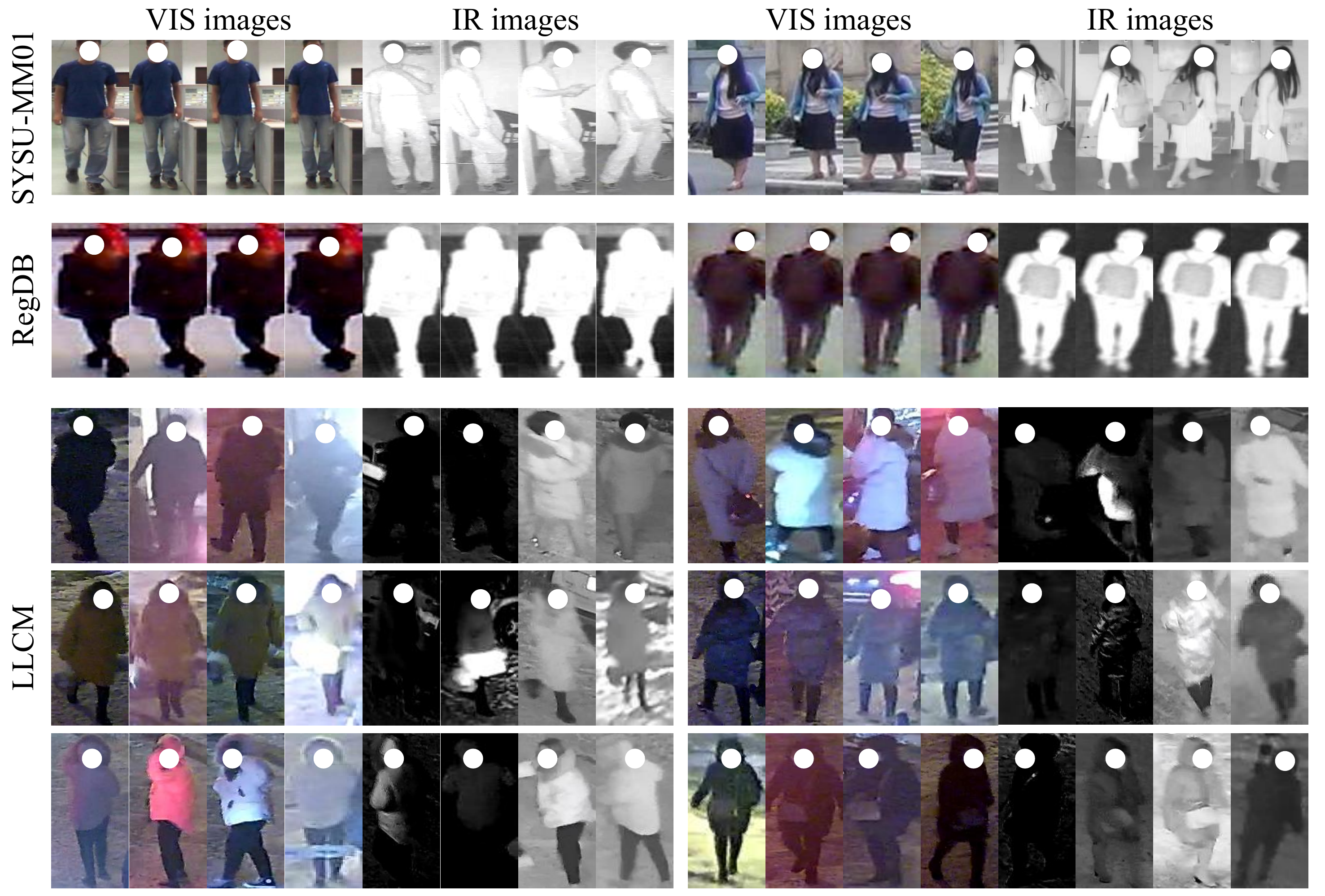}}
\vspace{-0.2cm}
\caption{Comparison of person images on the SYSU-MM01 (1st row), RegDB (2nd row), and LLCM (3rd-5th rows) datasets. Each row shows four VIS images and four IR images of two identities. It is obvious that our LLCM contains a more challenging and realistic VIReID environment.}
\vspace{-0.3cm}
\label{img:img2}
\end{figure}

In this paper, we propose a novel augmentation network in the embedding space for the VIReID task, called diverse embedding expansion network (DEEN), which consists of a diverse embedding expansion (DEE) module and a multistage feature aggregation (MFA) block. The proposed DEE module can generate more embeddings followed by a novel center-guided pair mining (CPM) loss to drive the DEE module to focus on learning with the diverse feature representations. As illustrated in Fig. \ref{img:img1}, by exploiting the generated embeddings with diverse information, the proposed DEE module can achieve the performance improvement by using more diverse embeddings. The proposed MFA block can aggregate the features from different stages for mining potential channel-wise and spatial feature representations, which increases the network’s capacity for mining different-level diverse embeddings. 

Moreover, we observe that the existing VIReID datasets are captured under the environments with sufficient illumination. %However, the VIReID task in real world has to deal with the challenges including low-light and low-illumination scenes. 
However, the performance of the VIReID methods may be seriously affected by drastic illumination changes or low illuminations. Therefore, we collect a challenging low-light cross-modality dataset, called LLCM dataset, which is shown in Fig. \ref{img:img2}. Compared with the other VIReID datasets, the LLCM dataset contains a larger number of identities and images captured under low-light scenes, which introduces more challenges to the real-world VIReID task.

In summary, the main contributions are as follows:

$\bullet$ We propose a novel diverse embedding expansion (DEE) module with a center-guided pair mining (CPM) loss to generate more embeddings for learning the diverse feature representations. We are the first to augment the embeddings in the embedding space in VIReID. Besides, we also propose an effective multistage feature aggregation (MFA) block to mine potential channel-wise and spatial feature representations. 
 
$\bullet$ With the incorporation of DEE, CPM loss and MFA into an end-to-end learning framework, we propose an effective diverse embedding expansion network (DEEN), which can effectively reduce the modality discrepancy between the VIS and IR images. 

$\bullet$ We collect a low-light cross-modality (LLCM) dataset, which contains 46,767 images of 1,064 identities captured under the environments with illumination changes and low illuminations. The LLCM dataset has more new and important features, which can facilitate the research of VIReID towards practical applications. 

$\bullet$ Extensive experiments show that the proposed DEEN outperforms the other state-of-the-art methods for the VIReID task on three challenging datasets. 

%$\bullet$ We propose a novel diverse embedding expansion (DEE) module with a center-guided pair mining (CPM) loss to generate diverse embeddings to learn the informative feature representations, which can effectively reduce the modality discrepancy between VIS and IR images. 

%$\bullet$ We propose an effective multistage feature aggregation (MFA) block to mine potential channel-wise and spatial feature representations. The proposed MFA block can greatly increase the network’s capacity for learning different-level diverse embeddings.

%$\bullet$ We collect a low-light cross-modality (LLCM) dataset, which contains 46,767 images of 1,064 identities captured  under the environments with illumination changes and low illuminations. The LLCM dataset has more new and important features, which can facilitate the research of VIReID towards practical applications. 

%$\bullet$ Extensive experiments show that the proposed method outperforms other state-of-the-art methods for the challenging VIReID task on the SYSU-MM01, RegDB and LLCM datasets. 
\section{Related Work}

Generally speaking, there are two main categories of methods in VIReID: the image-level methods and the feature-level methods. 

%The image-level VIReID methods attempt to exchange image styles across modes

The image-level VIReID methods try to transform one modality into the other for reducing the modality discrepancy between the VIS and IR images in the image space. For this purpose, some GANs-based \cite{wang2019learning, choi2020hi, wang2019aligngan, wang2020cross} methods are proposed to perform identity-preserving person image style transformation for aligning cross-modality images and alleviating the problem of limited data. These methods often design complex generative models to align cross-modality images. However, due to the lack of VIS-IR image pairs, the generated images are unavoidably accompanied by some noises. X-modality \cite{li2020infrared} and its variations \cite{zhang2021towards, wei2021syncretic} apply a lightweight network to introduce an auxiliary middle modality to assist the cross-modality search task. However, there is still a modality gap between this middle modality and the VIS / IR modality.

\begin{figure*}[t]
\centering
\includegraphics[height=7.6cm,width=16.5cm]{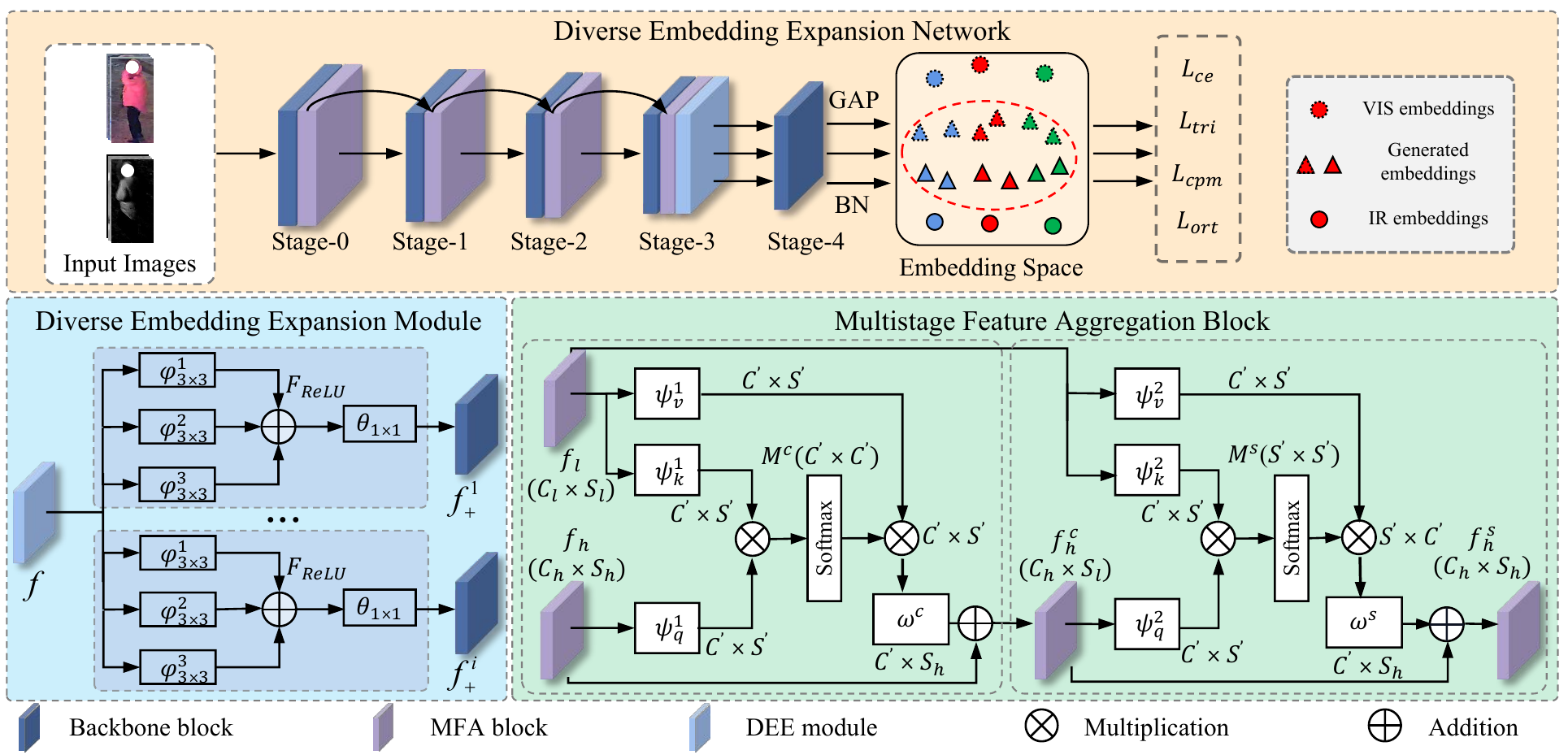}
\vspace{-0.2cm}
\caption{The pipeline of the proposed network, which includes a DEE module and a MFA block. The DEE module can generate more embeddings with a novel CPM loss to learn diverse feature representations. The MFA block can aggregate the embeddings from different stages for mining diverse channel-wise and spatial feature representations.}
\vspace{-0.3cm}
\label{img:img3}
\end{figure*}

The feature-level methods aim to find a modality-shared and modality-specific feature space, where the modality gaps can be minimized. For this purpose, MAUM \cite{LiuJialun_2022_CVPR} tries to learn cross-modality metrics in two uni-directions to further enhance them with memory-based augmentation. RFM \cite{tan2023exploring} introduces a cross-center loss to explore a more compact intra-class distribution.
%DG-VAE \cite{pu2020dual} adopts a variational auto-encoder to disentangle an identity-discriminable and an identity-ambiguous cross-modality feature subspace. 
%CM-NAS \cite{fu2021cm} introduces a BN-oriented search space, where the standard optimization can be implemented. 
%MCLNet \cite{hao2021cross} proposes to minimize the inter-modality discrepancy while maximizing the cross-modality similarity. 
DCLNet \cite{Sun_2022_ACM} encourages the positive pixels with the same semantic information to be close, while it simultaneously pushes the negative pixels away. cmGAN \cite{dai2018cross} designs a cutting-edge discriminator to learn discriminative representations from different modalities. %MPANet \cite{wu2021discover} introduces a modality alleviation module and a pattern alignment module to jointly extract discriminative features. 
However, the large modality gaps between the VIS and IR images make it difficult to project the cross-modality images into a common space directly \cite{gao2021mso, pu2020dual, tian2021farewell, miao2021modality}.

\section{Method}

%In this section, we elaborate on the proposed DEEN, a novel augmentation method in the embedding space for VIReID task. Specifically, in Section \ref{sub:sub1}, we present the overview of DEEN. Then, in Section \ref{sub:sub2}, we elaborate on the design of DEE module, followed by the CPML. Furthermore, in Section \ref{sub:sub3}, we present an effective feature aggregation strategy for leveraging different-level diverse potential features. Finally, in Section  \ref{sub:sub4}, we adopt a multi-loss strategy to jointly optimize the proposed method.
\subsection{Model Architecture}
\label{sub:sub1}

Fig. \ref{img:img3} provides an overview of the proposed diverse embedding expansion network (DEEN), which utilizes a two-stream ResNet-50 network \cite{he2016deep, ye2020dynamic} as the backbone. The VIS-IR features are fed into the proposed diverse embedding expansion (DEE) module to generate more embeddings. Then, a center-guided pair mining (CPM) loss is proposed to make the generated embeddings as diverse as possible for learning informative feature representations.
Besides, we incorporate an effective MFA block to aggregate the features from different stages for mining diverse channel-wise and spatial feature representations. %After the convolutional layers and global average pooling (GAP) layer, a batch normalization (BN) layer is used to make the loss easier to converge \cite{Luo2019Bags}. 
During the training stage, all the features before and after the batch normalization (BN) layer are fed into different losses to jointly optimize DEEN. 

\subsection{Diverse Embedding Expansion Module}
\label{sub:sub2}

%The proposed DEE module includes a diverse embedding generation (DEG) module and center-guided pair mining (CPM) loss.
%\subsubsection{Diverse Embedding Generation Module}
%Due to the natural difference between the reflectivity of the VIS spectrum and the emissivity of the IR spectrum, there is a large modality discrepancy between the images of these two different modalities. However, most of the existing methods cannot effectively reduce the modality difference due to insufficient training data. To this end, we propose to generate more embeddings to tackle this issue.

%For convenience, we first define the VIReID task. The proposed DEEN takes an image pair of the same identities but different modalities as the input. Let $\textbf{\textit{I}} = \left\{{\textbf{I}_{v}, \textbf{I}_{n}|\textbf{I}_{v}, \textbf{I}_{n} \in{\mathbb{R}}^{c \times h \times w} } \right\}$ denote the dataset consisting of the VIS images $\textbf{I}_v$ and the IR images $\textbf{I}_n$ , where $\textbf{\textit{I}}$ is a set of images, $c , h, w$ is the channel, hight and weight of $\textbf{I}_v$ and $\textbf{I}_n$, respectively. 
The proposed DEE module is used to generate more embeddings to alleviate the problem of insufficient training data by using a multi-branch convolutional generation structure. 
Specifically, for each branch of DEE, we firstly use three $3 \times 3$ dilated convolutional layers $\varphi^{1}_{3 \times 3}$, $\varphi^{2}_{3 \times 3}$, $\varphi^{3}_{3 \times 3}$ with different dilation ratios $(1, 2, 3)$ to reduce the number of feature maps $\textbf{f}$ to 1 / 4 of its own size, and then we obtain the feature maps by combining them into one feature map, followed by a ReLU activation layer $\textbf{F}_{ReLU}$ to improve the non-linear representation capability of the DEE. Then, another convolutional layer $\theta_{1 \times 1}$ with a kernel in size of $1 \times 1$ is applied to the obtained feature map to change its dimension as same as $\textbf{f}$. Thus, the generated embeddings $\textbf{f}^{i}_{+}$ of the $i$-th branch can be written as follows:
\begin{small}
\begin{equation}
\textbf{f}^{i}_{+} = \theta_{1 \times 1}(\textbf{F}_{ReLU}(\varphi^{1}_{3 \times 3}(\textbf{f}) + \varphi^{2}_{3 \times 3}(\textbf{f}) + \varphi^{3}_{3 \times 3}(\textbf{f}))).
\end{equation}
\end{small}

Then, all the generated embeddings are concatenated together and used as the input to the next stage of the backbone network.

\subsection{Center-Guided Pair Mining Loss}
As we can see from the above operation, the DEE module can only generate more embeddings using a multi-branch convolutional block. However, this operation cannot effectively obtain diverse embeddings. Thus, we apply the following three properties to constrain the generated embeddings as diverse as possible to effectively reduce the modality discrepancy between the VIS and IR images:

\noindent\textbf{Property 1: The generated embeddings should be as diverse as possible to effectively learn the informative feature representations.} This means that we need to push away the distances between the generated embeddings and the original embeddings to learn diverse features and mine diverse cross-modality clues. 

\noindent\textbf{Property 2: The generated embeddings should facilitate reducing the modality discrepancy between the VIS and IR images.} This means that we need to pull close the distances between the embeddings generated from the VIS modality and the original IR embeddings. Similarly, we also need to pull close the distances between the embeddings generated from the IR modality and the original VIS embeddings. 

\noindent\textbf{Property 3: The intra-class distance should be less than the inter-class one.} By \textbf{Property 2}, it pushes close the distance between the generated embeddings and the original ones, which may cause the embeddings of different classes to become close. Thus, it is necessary to keep the intra-class distance less than the inter-class distance.

\begin{figure}[t]
\centerline{\includegraphics[height=3.6cm,width=7.8cm]{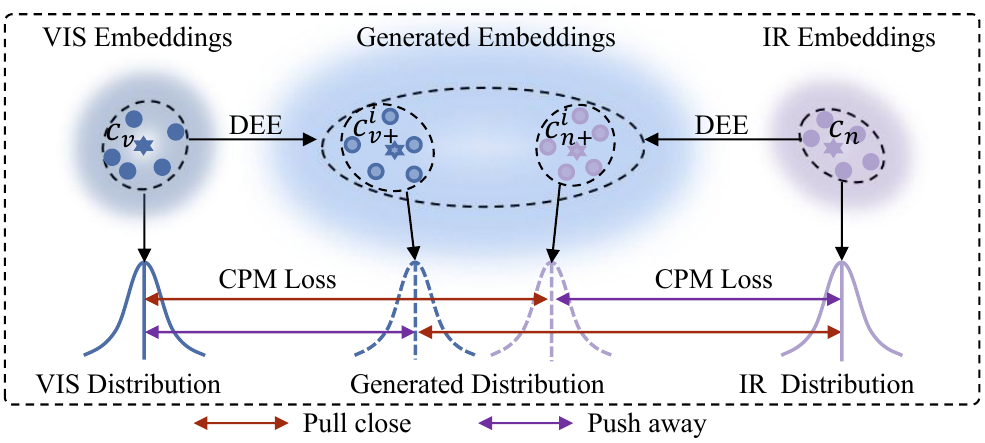}}
\vspace{-0.2cm}
\caption{Illustration of the proposed CPM loss for DEE.}
\vspace{-0.3cm}
\label{img:img4}
\end{figure}

As shown in Fig. \ref{img:img4}, for embeddings generated from the VIS modality, the CPM loss can be formulated as follows:
\begin{small}
\begin{equation}
\label{eq2}
\begin{split}
{\mathcal{L}(\textbf{f}_v, \textbf{f}_n, \textbf{f}_{v+}^{i})} = [\textbf{\textit{D}}(\textbf{f}^{j}_{n}, \textbf{f}^{i, j}_{v+}) - \textbf{\textit{D}}(\textbf{f}^{j}_{v}, \textbf{f}^{i, j}_{v+}) - \textbf{\textit{D}}(\textbf{f}^{j}_{v}, \textbf{f}^{k}_{v})]_{+},
\end{split}
\end{equation}
\end{small}where ${\textbf{\textit{D}}}(\cdot, \cdot)$ is the Euclidean distance between two embeddings. $\textbf{f}_{v}$ and $\textbf{f}_{n}$ are the original embeddings from the VIS and IR modalities, and $\textbf{f}^{i}_{v+}$ is the embeddings generated from the $i$-th branch of the VIS modality. $j,k$ are different identities in a minibatch and $[z]_{+} = max(z,0)$. In Eq. (\ref{eq2}), the first term can pull the generated embeddings $\textbf{f}^{i}_{v+}$ towards the original IR's embeddings $\textbf{f}_{n}$ to reduce the modality discrepancy between $\textbf{f}^{i, j}_{v+}$ and $ \textbf{f}^{j}_n$. The second term can push the generated embeddings $\textbf{f}_{v+}$ away from the VIS's embeddings $\textbf{f}_{v}$ to enable $\textbf{f}_{v+}$ to learn informative feature representations. The third term can make the intra-class distance less than the inter-class distance.

Then, we use the embedding centers $\textbf{c}_{v}$ and $\textbf{c}_{n}$ of each class to make the centers of generated embeddings $\textbf{c}^{i}_{v+}$ and $\textbf{c}^{i}_{n+}$ more discriminative, and introduce a margin term $\alpha$ to balance the three terms in Eq. (\ref{eq2}). Thus, for the embeddings from VIS, the CPM loss is formulated as follows:
\begin{small}
\begin{equation}
\label{eq3}
\begin{split}
{\mathcal{L}(\textbf{c}_v, \textbf{c}_n, \textbf{c}_{v+}^{i})} = [\textbf{\textit{D}}(\textbf{c}^{j}_{n}, \textbf{c}^{i, j}_{v+}) - \textbf{\textit{D}}(\textbf{c}^{j}_{v}, \textbf{c}^{i, j}_{v+}) - \textbf{\textit{D}}(\textbf{c}^{j}_{v}, \textbf{c}^{k}_{v}) + \alpha]_{+}.
\end{split}
\end{equation}
\end{small}
%where $\textbf{c}_{v}$ and $\textbf{c}_{n}$ are the centers of each class from the VIS and IR modalities. $\textbf{c}^{i}_{v+}$ is the class centers of embeddings generated from the $i$-th branch of the VIS modality. 
Similarly, for the class centers $\textbf{c}^{i}_{n+}$ of embeddings generated from IR, we have:
\begin{small}
\begin{equation}
\label{eq4}
\begin{split}
{\mathcal{L}(\textbf{c}_v, \textbf{c}_n, \textbf{c}_{n+}^{i})} = [\textbf{\textit{D}}(\textbf{c}^{j}_{v}, \textbf{c}^{i, j}_{n+}) - \textbf{\textit{D}}(\textbf{c}^{j}_{n}, \textbf{c}^{i, j}_{n+}) - \textbf{\textit{D}}(\textbf{c}^{j}_{n}, \textbf{c}^{k}_{n}) + \alpha]_{+}.
\end{split}
\end{equation}
\end{small}

Thus, the final CPM loss can be formulated as follows:
\begin{small}
\begin{equation}
\label{eq5}
\mathcal{L}_{cpm} = {\mathcal{L}(\textbf{c}_v, \textbf{c}_n, \textbf{c}_{v+}^{i})} + {\mathcal{L}(\textbf{c}_v, \textbf{c}_n, \textbf{c}_{n+}^{i})}.
\end{equation}
\end{small}

Besides, to ensure that the generated embeddings from different branches can capture different informative feature representations, we force these different embeddings generated by different branches orthogonal to minimize the overlapping elements. Therefore, the orthogonal loss can be formulated as follows:
\begin{small}
\begin{equation}
\label{eq6}
\mathcal{L}_{ort} = \sum\limits_{m=1}^{i-1} \sum\limits_{n=m + 1}^i({\textbf{f}^{m}_{+}}^{T}\textbf{f}^{n}_{+}),
\end{equation}
\end{small}where $m$ and $n$ are the $m$-th and $n$-th generated embeddings from the original embeddings, respectively. The orthogonal loss can enforce the generated embeddings to learn more informative feature representations.

\subsection{Multistage Feature Aggregation Block}
\label{sub:sub3}

Features aggregation of different levels has been demonstrated to be helpful to semantic segmentation, classification and detection task\cite{Chen_2020_CVPR, zhu2019asymmetric, Zhou_2019_ICCV}. To aggregate the features from different stages for mining diverse channel-wise and spatial feature representations, we incorporate an effective channel-spatial multistage feature aggregation (MFA) block to aggregate multi-stage features inspired by \cite{wang2018non}. 

Next, we elaborate on the detail of the MFA block, which is shown in Fig. \ref{img:img3}. 
Specifically, we consider two types of source features for the channel-spatial aggregation block in each stage of the backbone network: low-level feature maps $\textbf{f}_l\in\mathbb{R}^{C_l \times H_l \times W_l}$ before the stage and  high-level feature maps $\textbf{f}_h\in\mathbb{R}^{C_h \times H_h \times W_h}$ after the stage, where C, W and H denote the number of the channel, width and height of features, respectively. First, we employ three 1×1 convolutional layers $\psi^{1}_{q}, \psi^{1}_{v}, \psi^{1}_{k}$ to transform $\textbf{f}$ into three compact embeddings: $\psi_{q}^{1}(\textbf{f}_{h})$ %$\in\mathbb{R}^{C^{'} \times S^{'}}$
, $\psi_{v}^{1}(\textbf{f}_{l})$ %$\in\mathbb{R}^{C^{'} \times S^{'}}$
and $\psi_{k}^{1}(\textbf{f}_{l})$ %$\in\mathbb{R}^{C^{'} \times S^{'}}$
.
Then, we compute the channel-wise similarity matrix $\textbf{M}^{c} \in\mathbb{R}^{C^{'} \times C^{'}}$ by matrix multiplication followed by softmax:
\begin{small}
\begin{equation}
\textbf{M}^{c} = \textbf{F}_{softmax}(\psi_{q}^{1}(\textbf{f}_h) \times \psi_{k}^{1}(\textbf{f}_l)).
\end{equation}
\end{small}

Consequently, we implement the channel-wise multistage feature aggregation by restoring the channel dimension by the matrix multiplication of $\psi_{v}^{1}(\textbf{f}_l)$ and $\textbf{M}^{c}$. After that, another $1 \times 1$ convolutional layer $\omega^{c}$ is applied to transform the size of the above feature maps to that of $\textbf{f}_h$. Finally, we get the output by adding $\textbf{f}_{h}$ to it by matrix addition:
\begin{small}
\begin{equation}
\textbf{f}_{h}^{c} = {{\omega^{c}} (\psi_{v}^{1}(\textbf{f}_l) \times \textbf{M}^{c}) + \textbf{f}_{h}}.
\end{equation}
\end{small}
After that, $\textbf{f}_{h}^{c}$ obtained from the above operations and the low-level feature map $\textbf{f}_l$ are used to perform the spatial feature aggregation operation, which is similar to the channel-wise multistage feature aggregation operation. Finally, we get the MFA’s output as follows:
\begin{small}
\begin{equation}
\textbf{f}_{h}^{s} = {{\omega^{s}}(\psi_{v}^{2}(\textbf{f}_l) \times \textbf{M}^{s}) + \textbf{f}_{h}^{c}},
\end{equation}
\end{small}
where $\omega^{s}$ and $\psi_{v}^{2}$ are two $1 \times 1$ convolutional layers, and $\textbf{M}^{s}$ is the spatial similarity matrix.

\subsection{Multi-Loss Optimization}
\label{sub:sub4}

%The features output by DEE are fed into the proposed DEEN to assist the optimization of the network. 
Besides the proposed ${\mathcal{L}_{cpm}}$ and ${\mathcal{L}_{ort}}$, we also combine the cross-entropy loss ${\mathcal{L}_{ce}}$ \cite{Luo2019Bags} and the triplet loss ${\mathcal{L}_{tri}}$ \cite{hermans2017defense} to jointly optimize the network in an end-to-end manner by minimizing the sum of these four losses ${\mathcal{L}_{total}}$, which can be formulated as follows:
\begin{small}
\begin{equation}
\label{eq11}
{\mathcal{L}_{total}} = {\mathcal{L}_{ce}} + {\mathcal{L}_{tri}} + \lambda_1{\mathcal{L}_{cpm}} + \lambda_2{\mathcal{L}_{ort}},
\end{equation}
\end{small}where $\lambda_1$ and $\lambda_2$ are the coefficients to control the relative importance of the loss terms.

\section{LLCM Dataset}
\label{sub:sub5}

\subsection{Dataset Description}

\begin{figure}[t]\footnotesize
\centerline{\includegraphics[height=5cm,width=8cm]{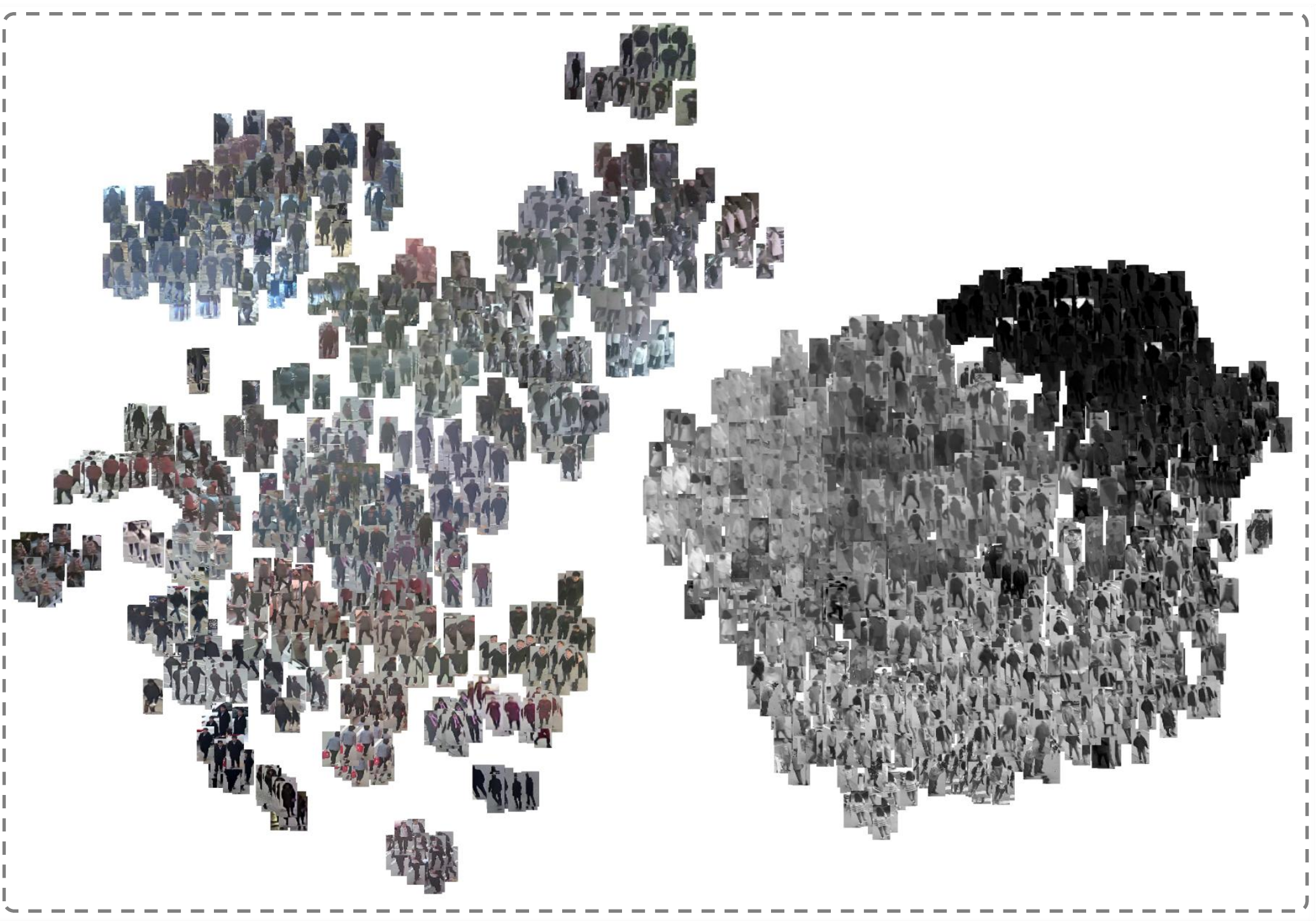}}
\vspace{-0.2cm}
\caption{The distribution of the LLCM's images in the 2D space. It can be seen that the images under different light conditions present different styles, which further increases the modality discrepancy between the VIS and IR images.}
\vspace{-0.3cm}
\label{img:img5}
\end{figure}

In this paper, we collect a new challenging low-light cross-modality dataset, called LLCM dataset. The LLCM dataset utilizes a 9-camera network deployed in low-light environments, which can capture the VIS images in daytime and capture the IR images at night. %We select 100 days from January to April for the video collection. For each day, the video is collected from dusk to night, covering different low-light and no-light environments. Then, YOLOv3 \cite{redmon2018yolov3} is adopted for person detection, which is fast to detect more than 300 hours of raw video for our dataset. 
For protecting the personal privacy information, we utilize MTCNN \cite{zhang2016joint} to get the bounding boxes of persons' faces and blur those regions. %Totally, the completed dataset contains 46,767 images of 1,064 identities. 
We make sure that each annotated identity is captured by both the VIS and IR cameras. Some examples from the LLCM dataset are shown in Fig. \ref{img:img2}. As shown in Tab. \ref{tab1:tab1}, compared with the existing VIReID datasets, the LLCM dataset has the following new and important features:

First, %the LLCM dataset contains severe illumination changes. the images in the LLCM are captured under complex low-light environment and they are in both the VIS and IR modalities, which is a common problem in the real scenes. 
the images in the LLCM dataset are captured under complex low-light environment for both the VIS and IR modalities, which contains severe illumination changes and is a common problem in the real scenes. As  Fig. \ref{img:img2} and Fig. \ref{img:img5} shown, the severe light conditions can change the color of persons' clothes and cause the loss of texture information of the clothes, which introduces great challenges to VIReID. 
Second, the LLCM dataset has a larger number of identities and bounding boxes. This dataset contains 46,767 bounding boxes of 1,064 identities, making it the largest VIReID dataset at present (see Tab \ref{tab:tab1}). 
Third, the LLCM dataset is collected in over 100 days from January to April, and different climate conditions and cloth styles are considered. Long-term data collection helps to study the VIReID task in different climates and clothing styles, which increases the generalization of the VIReID model. 

Besides, considering the real-world applications, the LLCM dataset also contains many images that suffer from various challenges, such as motion bluring, pose variation, camera view changes, occlusion, low resolution and others. All in all, the LLCM dataset is a challenging dataset for the VIReID task, which can further facilitate the research of VIReID towards practical applications.

\subsection{Evaluation Protocol}

\begin{table}
\centering
  \renewcommand{\arraystretch}{0.8}
\fontsize{7.3pt}{13pt}\selectfont
\begin{tabular}{l|cccccc}
\hline
Datasets                     & IDs   & Images & VIS / IR cam. & low-light \\
\hline
RegDB \cite{nguyen2017person}& 412   & 8,240  & 1 / 1        & \XSolidBrush    \\   
SYSU-MM01 \cite{wu2017rgb}   & 491   & 38,271 & 4 / 2       & \XSolidBrush   \\   
LLCM                 & \textbf{1,064} & \textbf{46,767} & \textbf{9 / 9} & \CheckmarkBold  \\ \hline
\end{tabular}%}
  \vspace{-0.2cm}
\caption{Comparison between the LLCM and other two popular VIReID datasets.}\label{tab1:tab1}
  \vspace{-0.5cm}
  \label{tab:tab1}
\end{table}

We divide the LLCM dataset into a training set and a testing set at a ratio about 2:1. The training set contains 30,921 bounding boxes of 713 identities (16,946 bounding boxes are from the VIS modality and 13,975 bounding boxes are from the IR modality), and the testing set contains 13,909 bounding boxes of 351 identities (8,680 bounding boxes are from the VIS modality and 7,166 bounding boxes are from the IR modality). Similar to the RegDB \cite{nguyen2017person} dataset, both the VIS to IR mode and the IR to VIS mode are used to evaluate the performance of the VIReID models. During the testing stage, for each camera, we randomly choose one image from the images of each identity to form the gallery set for evaluation the performance of the models. We repeat the above evaluation 10 times with random split of the gallery set and report the average performance.

\begin{table*}\footnotesize
  \centering
  \renewcommand{\arraystretch}{1.0}
  \resizebox{\textwidth}{!}{
  \begin{tabular}{l|cccc|cccc|cccc|cccc}
  %\toprule[1pt]
  \hline
   \multirow{3}{*}{Methods} & \multicolumn{8}{c|}{SYSU-MM01} & \multicolumn{8}{c}{RegDB} \\
   \cline{2-17}
   &\multicolumn{4}{c|}{All Search} &\multicolumn{4}{c|}{Indoor Search}&\multicolumn{4}{c|}{VIS to IR} &\multicolumn{4}{c}{IR to VIS} \\
   \cline{2-17}
   & R-1 & R-10 & R-20 & mAP   & R-1 & R-10 & R-20 & mAP & R-1 & R-10 & R-20 & mAP & R-1 & R-10 & R-20 & mAP\\
   \hline
    BDTR\cite{ye2018visible}           & 17.0 & 55.4 & 72.0 & 19.7 & -    & -    & -    & -       & 33.6 & 58.6 & 67.4 & 32.8 & 32.9 & 58.5 & 68.4 & 32.0 \\
    D$^{2}$RL\cite{wang2019learning}   & 28.9 & 70.6 & 82.4 & 29.2 & -    & -    & -    & -       & 43.4 & 66.1 & 76.3 & 44.1 & -    & -    & -    & -       \\
    Hi-CMD\cite{choi2020hi}            & 34.9 & 77.6 & -    & 35.9 & -    & -    & -    & -       & 70.9 & 86.4 & -    & 66.0 & -    & -    & -    & -       \\
    JSIA-ReID\cite{wang2020cross}      & 38.1 & 80.7 & 89.9 & 36.9 & 43.8 & 86.2 & 94.2 & 52.9    & 48.1 & -    & -    & 48.9 & 48.5 & -    & -    & 49.3    \\
    AlignGAN\cite{wang2019aligngan}    & 42.4 & 85.0 & 93.7 & 40.7 & 45.9 & 87.6 & 94.4 & 54.3    & 57.9 & -    & -    & 53.6 & 56.3 & -    & -    & 53.4    \\
    X-Modality\cite{li2020infrared}    & 49.9 & 89.8 & 96.0 & 50.7 & -    & -    & -    & -       & 62.2 & 83.1 & 91.7 & 60.2 & -    & -    & -    & -       \\
    DDAG\cite{ye2020dynamic}           & 54.8 & 90.4 & 95.8 & 53.0 & 61.0 & 94.1 & 98.4 & 68.0   & 69.3 & 86.2 & 91.5 & 63.5 & 68.1 & 85.2 & 90.3 & 61.8    \\
    LbA\cite{park2021learning}         & 55.4 & -    & -    & 54.1 & 58.5 & -    & -    & 66.3    & 74.2 & -    & -    & 67.6 & 67.5 & -    & -    & 72.4    \\
    NFS\cite{chen2021neural}           & 56.9 & 91.3 & 96.5 & 55.5 & 62.8 & 96.5 & 99.1 & 69.8  & 80.5 & 91.6 & 95.1 & 72.1 & 78.0 & 90.5 & 93.6 & 69.8    \\
    CM-NAS\cite{fu2021cm}              & 60.8 & 92.1 & 96.8 & 58.9 & 68.0 & 94.8 & 97.9 & 52.4    & 82.8 & 95.1 & 97.7 & 79.3 & 81.7 & 94.1 & 96.9 & 77.6    \\
    MCLNet\cite{hao2021cross}          & 65.4 & 93.3 & 97.1 & 62.0 & 72.6 & 97.0 & 99.2 & 76.6    & 80.3 & 92.7 & 96.0 & 73.1 & 75.9 & 90.9 & 94.6 & 69.5    \\
    FMCNet\cite{Zhang_2022_CVPR}       & 66.3 & -    & - & 62.5 & 68.2 & -    & - & 74.1 & 89.1 & -    & - & 84.4 & 88.4 & -    & - & 83.9 \\
    SMCL\cite{wei2021syncretic}        & 67.4 & 92.9 & 96.8 & 61.8 & 68.8 & 96.6 & 98.8 & 75.6    & 83.9 & -    & -    & 79.8 & 83.1 & -    & -    &78.6     \\
    DART\cite{Yang_2022_CVPR}          & 68.7 & \textcolor{blue}{96.4} & \textcolor{blue}{99.0} & 66.3 & 72.5 & 97.8 & 99.5 & 78.2& 83.6 & -    & -    & 75.7 & 82.0 & -    & -    & 73.8    \\
    CAJ\cite{ye2021channel}            & 69.9 & 95.7 & 98.5 & 66.9 & 76.3 & 97.9 & 99.5 & 80.4  & 85.0 & 95.5 & \textcolor{blue}{97.5} & 79.1 & 84.8 & 95.3 & 97.5 & 77.8 \\
    MPANet\cite{wu2021discover}        & 70.6 & 96.2 & 98.8 & 68.2 & 76.7 & \textcolor{blue}{98.2} &  \textcolor{blue}{99.6} & 81.0 & 82.8 & -    & -    & 80.7 & 83.7 & -    & -    & 80.9    \\
    MMN \cite{zhang2021towards}        & 70.6 & 96.2 & \textcolor{blue}{99.0} & 66.9 & 76.2 & 97.2 & 99.3 & 79.6   & \textcolor{red}{\textbf{91.6}} & \textcolor{blue}{97.7} & \textcolor{red}{\textbf{98.9}} & \textcolor{blue}{84.1} & 87.5 & \textcolor{blue}{96.0} & \textcolor{blue}{98.1} & 80.5 \\ 
    DCLNet\cite{Sun_2022_ACM}          & 70.8 & -    & -    & 65.3 & 73.5 & -    & -    & 76.8    & 81.2 & -    & -    & 74.3 & 78.0 & -    & -    & 70.6    \\
    MAUM\cite{LiuJialun_2022_CVPR} & \textcolor{blue}{71.7}  & -    & - & \textcolor{blue}{68.8}  & \textcolor{blue}{77.0}  & -    & - & \textcolor{blue}{81.9} & 87.9 & - & - & \textcolor{red}{\textbf{85.1}} & \textcolor{blue}{87.0} & - & -& \textcolor{red}{\textbf{84.3}}   \\
  \hline     
    DEEN (ours)              & \textcolor{red}{\textbf{74.7}} & \textcolor{red}{\textbf{97.6}} & \textcolor{red}{\textbf{99.2}} & \textcolor{red}{\textbf{71.8}} & \textcolor{red}{\textbf{80.3}} & \textcolor{red}{\textbf{99.0}} & \textcolor{red}{\textbf{99.8}} & \textcolor{red}{\textbf{83.3}}  & \textcolor{blue}{91.1} & \textcolor{red}{\textbf{97.8}} & \textcolor{red}{\textbf{98.9}} & \textcolor{red}{\textbf{85.1}} & \textcolor{red}{\textbf{89.5}} & \textcolor{red}{\textbf{96.8}} & \textcolor{red}{\textbf{98.4}} & \textcolor{blue}{83.4}   \\
   
    \hline
    %\toprule[0.pt]
    \end{tabular}}
    \vspace{-0.2cm}
    \caption{Comparisons between the proposed DEEN and some state-of-the-art methods on the SYSU-MM01 and RegDB datasets.
    %Performance obtained by the competing methods on the RegDB and SYSU-MM01 datasets. %R-1, 10, 20 denotes the Rank-1, 10, 20 accuracy, respectively. %Here, all the results obtained by the competing methods are meatured using the single-query rule.
    %The bold font denotes the best performance and the underline represents the second best performance. 
    %For fair comparison, we divide the compared methods into two groups:  the feature-based methods (top row) and the image-based methods (bottom row).}%The ``†'' represents feature extraction based methods, the ``*'' represents image generation based methods and the ``§'' represents a lightweight network to generate images.
    }
        \label{tab:tab2}
        \vspace{-0.3cm}
\end{table*}

\section{Experiments}

%In this section, we compare the proposed DEEN with several state-of-the-art methods and conduct ablation studies to analyze the key components in DEEN.
\subsection{Datasets}

The SYSU-MM01 dataset \cite{wu2017rgb} contains 491 identities captured by 4 VIS cameras and 2 IR cameras, %. The training set contains $19,659$ VIS images and $12,792$ IR images of $395$ identities, and the testing set contains $3,803$ IR images of $96$ identities as the query set. The gallery set is determined by the testing mode, 
including the All-Search and Indoor-Search modes. For the All-Search mode, all the images captured by all the VIS cameras are used as the gallery set. For the Indoor-Search mode, only the images captured by two indoor VIS cameras are used as the gallery set. The RegDB dataset \cite{nguyen2017person} consists of 412 identities, and each identity has 10 VIS images and 10 IR images captured by a pair of overlapping cameras. %Following the evaluation protocol of \cite{ye2018hierarchical}, we evaluate the competing methods in both VIS to IR and IR to VIS modes. Following the evaluation protocol of \cite{ye2018hierarchical}, we repeat 10 trails with a random half-half split of the dataset: half of identities are used for training and the other half are used for testing. %The final results are based on an average of 10 times testing.

%\textbf{Evaluation Protocol}. The standard Cumulative Matching Characteristics (CMC) curve and the mean Average Precision (mAP) are used as the performance evaluation metrics in our experiments. 

\subsection{Implementation Details}
%The backbone network employed in the proposed DEEN is first pre-trained on ImageNet \cite{Deng2009ImageNet}, and then finetuned on the training images. 
All the input images are firstly resized to $3 \times 384 \times 144$, and the random horizontal flip and random erasing \cite{zhong2020random} techniques are adopted during the training phase. The initial learning rate is set to $1\times10^{-2}$ and then it increases to $1\times10^{-1}$ after 10 epochs with a warm-up strategy. After that, we decay the learning rate to $1\times10^{-2}$ at 20 epoch, and further decay to $1\times10^{-3}$ and  $1\times10^{-4}$ at epoch 60 and epoch 120, respectively, until a total of 150 epochs. In each mini-batch, we randomly select 4 VIS images and 4 IR images of 6 identities for training. The SGD optimizer is adopted for training, where the momentum is set to 0.9. For the RegDB dataset, we remove stage-4 and plug the proposed DEE module into the DEEN after stage-2.%For the margin parameter in the CPML, we experimentally set it to 0.2. For the coefficients $\lambda_1$ and $\lambda_2$ in Eq. (\ref{eq1}), we set it to 1.2 and 0.01.

\subsection{Comparison with State-of-the-art Methods}

We firstly compare the proposed DEEN with several state-of-the-art methods to demonstrate the superiority of our method. The experimental results on the SYSU-MM01 and RegDB datasets are reported in Tab. \ref{tab:tab2}, and the results on our LLCM dataset are reported in Tab. \ref{tab:tab3}.

\textbf{SYSU-MM01 and RegDB:} From Tab. \ref{tab:tab2}, we can see that the results on the two datasets show that the proposed DEEN achieves the best performance against all other state-of-the-art methods. Specifically, for the All-Search mode on SYSU-MM01, DEEN achieves 74.7\% Rank-1 accuracy and 71.8\% mAP. For the Indoor-Search mode, DEEN achieves 80.3\% Rank-1 accuracy and 83.3\% mAP. 
For the VIS to IR mode on RegDB, DEEN achieves 91.1\% Rank-1 accuracy and 85.1\% mAP. %The proposed DEEN outperforms the second best MMN \cite{zhang2021towards} by 3.5\% Rank-1 accuracy and 5.1\% mAP, respectively. 
For the IR to VIS mode, the proposed DEEN also obtains 89.5\% Rank-1 accuracy and 83.4\% mAP. %The proposed DEEN outperforms the second best method FMCNet\cite{Zhang_2022_CVPR} by 4.2\% in Rank-1 accuracy and 2.6\% in mAP, respectively. 
The results validate the effectiveness of our method. Moreover, the results also demonstrate that the proposed DEEN can effectively reduce the modality discrepancy between the VIS and IR modalities.

\begin{table}\footnotesize
  \centering
  \tabcolsep=0.15cm
  \renewcommand{\arraystretch}{1.0}
  \fontsize{7.8pt}{9pt}\selectfont
  \begin{tabular}{l|cccc|cccc}
  %\toprule[1pt]
  \hline
   \multirow{3}{*}{Model} & \multicolumn{8}{c}{LLCM} \\
   \cline{2-9}
   &\multicolumn{4}{c|}{IR to VIS} &\multicolumn{4}{c}{VIS to IR}\\
   \cline{2-9}
   & R-1 & R-10 & R-20 & mAP   & R-1 & R-10 & R-20 & mAP \\
   \hline
    DDAG\cite{ye2020dynamic}      & 40.3 & 71.4 & 79.6 & 48.4 & 48.0 & 79.2 & 86.1 & 52.3    \\
    DDAG*\cite{ye2020dynamic}     & 41.0 & 73.4 & 81.9 & 49.6 & 48.5 & 81.0 & 87.8 & 53.0    \\
    AGW\cite{ye2020deep}          & 43.6 & 74.6 & 82.4 & 51.8 & 51.5 & 81.5 & 87.9 & 55.3 \\
    LbA\cite{park2021learning}       & 43.8 & 78.2 & 86.6 & 53.1 & 50.8 & 84.3 & 91.1 & 55.6 \\
    LbA*\cite{park2021learning}      & 44.6 & 78.2 & 86.8 & 53.8 & 50.8 & 84.6 & 91.1 & 55.9 \\
    %TSR\cite{ye2018visible}          & 45.4 & 76.0 & 83.5 & 53.6 & 53.4 & 83.0 & 89.0 & 57.0 \\
    AGW*\cite{ye2020deep}         & 46.4 & 77.8 & 85.2 & 54.8 & 56.0 & 84.9 & 90.6 & 59.1 \\
    CAJ\cite{ye2021channel}       & 48.8 & 79.5 & 85.3 & 56.6 & 56.5 & 85.3 & 90.9 & 59.8 \\
    DART\cite{Yang_2022_CVPR}            & 52.2 & 80.7 & 87.0 & \textcolor{blue}{59.8} & \textcolor{blue}{60.4} & 87.1 & 91.9 & \textcolor{blue}{63.2}     \\
    MMN\cite{zhang2021towards}        & \textcolor{blue}{52.5} & \textcolor{blue}{81.6} & \textcolor{blue}{88.4} &58.9 & 59.9 & \textcolor{blue}{88.5} & \textcolor{blue}{93.6} & 62.7 \\
    \hline
     
    DEEN (ours) & \textcolor{red}{\textbf{54.9}} & \textcolor{red}{\textbf{84.9}} & \textcolor{red}{\textbf{90.9}} & \textcolor{red}{\textbf{62.9}} & \textcolor{red}{\textbf{62.5}} & \textcolor{red}{\textbf{90.3}} & \textcolor{red}{\textbf{94.7}} & \textcolor{red}{\textbf{65.8}} \\
   
    \hline
    %\toprule[1pt]
    \end{tabular}%}
        \vspace{-0.2cm}
            \caption{Performance obtained by the competing methods on our LLCM dataset. The symbol of ``*'' represents the methods that we reproduced with the random erasing technique.}    \label{tab:tab3}
            \vspace{-0.3cm}
\end{table}

\textbf{LLCM:} Tab. \ref{tab:tab3} shows the results on our LLCM dataset. Here, we use several representative open-source methods to evaluate our LLCM dataset and compare them with our method. From Tab. \ref{tab:tab3} we can draw the following conclusions: the best method only obtains 54.9\% Rank-1 accuracy and 62.9\% mAP under the IR to VIS mode. The results of the existing methods on our LLCM dataset are generally unsatisfactory. This shows that, on one hand, our LLCM dataset is a very challenging dataset. On the other hand, the change of light has serious influence on the VIReID model. Besides, the proposed DEEN achieves the best performance under both the VIS to IR mode and the IR to VIS mode, which demonstrates the effectiveness of the proposed DEEN to reduce the modality gaps between the VIS and IR images.

\subsection{Ablation Studies} 

\textbf{Effectiveness of each component:} To evaluate the contribution of each component to DEEN, we conduct some ablation studies on the LLCM and SYSU-MM01 datasets by removing certain modules from DEEN and evaluate the influence on the performance. 
The overall settings remain the same, while only the module under evaluation is  used in or removed from DEEN. 
As shown in Tab. \ref{tab:tab4}, although the DEE module can generate more embeddings using a multi-branch convolutional block, which slightly improves the performance of the baseline, the results are not satisfactory. After being constrained by the proposed CPM loss to generate diverse embeddings, DEE can greatly improve the performance of the model and effectively reduce the modality discrepancy between the VIS and IR images. Besides, the proposed MFA block can improve the performance of the baseline by aggregating the features from different stages for mining diverse channel-wise and spatial feature representations. With the incorporation of DEE, CPM and MFA into an end-to-end learning framework, DEEN achieves an impressive performance improvement on two challenging VIReID datasets, which shows DEE and MFA can benefit from each other for generating diverse embeddings.

\begin{table}[t]\footnotesize
  \centering
  \renewcommand{\arraystretch}{0.9}
 \fontsize{7.8pt}{9pt}\selectfont
 %\resizebox{\textwidth}{!}{
  \begin{tabular}{lcccc|cc|cc}\hline%\toprule[1pt]
    & \multicolumn{4}{c|}{Settings} & \multicolumn{2}{c|}{LLCM} & \multicolumn{2}{c}{SYSU-MM01} \\
    \cline{1-9}
                            & DEE & $\mathcal{L}_{cpm}$  & $\mathcal{L}_{ort}$ & MFA        & R-1 & mAP & R-1 & mAP  \\
   \hline 
 &  &   &  &                       & 45.4 & 53.6  & 60.7	& 57.7   \\ 
& \Checkmark &   &     &           & 50.5 & 59.0  & 64.7 & 62.0  \\ 
&\Checkmark  & \Checkmark  &  &    & 53.1 & 61.1  & 69.2 & 66.2	 \\ 
& \Checkmark &  & \Checkmark &     & 51.5 & 60.1  & 65.3 & 63.2  \\
&\Checkmark  & \Checkmark & \Checkmark &  & 53.9 & 62.3  & 69.8	& 66.7  \\
 &  &  &  & \Checkmark             & 51.2 & 59.6  & 64.7	& 62.0 \\
&\Checkmark  & \Checkmark  &\Checkmark&\Checkmark & \textcolor{red}{\textbf{54.9}} & \textcolor{red}{\textbf{62.9}}  & \textcolor{red}{\textbf{74.7}} & \textcolor{red}{\textbf{71.8}} \\ 
    \hline
    %\toprule[1pt]
    \end{tabular}%}
    \vspace{-0.2cm}
      \caption{The influence of each component on the performance of the proposed DEEN.}  \label{tab:tab4}
      \vspace{-0.3cm}
\end{table}

\begin{table}[t]\footnotesize
  \centering
  \renewcommand{\arraystretch}{1.0}
   \fontsize{7.8pt}{9pt}\selectfont
  \begin{tabular}{c|cc|cc}\hline%\toprule[1pt]
    \multirow{2}{*}{Methods} & \multicolumn{2}{c|}{LLCM}& \multicolumn{2}{c}{SYSU-MM01} \\    \cline{2-5}
                         & R-1 & mAP & R-1 & mAP  \\
   \hline                   
%baseline                    & 45.4 	& 53.6  & 60.7	& 57.7 \\ \hline
DEE after stage-0	        & 48.5	& 57.1  & 63.4 & 59.4\\ 
DEE after stage-1           & 49.4 & 57.8  & 63.7	& 60.8\\
DEE after stage-2	        & 49.6 & 57.9  & 65.3	& 61.7 \\ 
DEE after stage-3	        & \textcolor{red}{\textbf{53.9}} & \textcolor{red}{\textbf{62.3}}   & \textcolor{red}{\textbf{69.8}}	& \textcolor{red}{\textbf{66.7}} \\ 
DEE after stage-4	        & 50.9	& 59.6  & 60.0	& 58.0\\ 
    \hline%\toprule[1pt]
    \end{tabular}
\vspace{-0.2cm}
  \caption{The influence of which stage of ResNet-50 to plug the DEE module.}  \label{tab:tab5}
  \vspace{-0.3cm}
\end{table}

\begin{table}[t]\footnotesize
  \centering
  \renewcommand{\arraystretch}{0.9}
   \fontsize{7.3pt}{10pt}\selectfont
  \begin{tabular}{l|cc|cc}\hline%\toprule[1pt]
    \multirow{2}{*}{Methods} & \multicolumn{2}{c|}{LLCM}& \multicolumn{2}{c}{SYSU-MM01} \\    \cline{2-5}
                         & R-1 & mAP & R-1 & mAP  \\
   \hline
%baseline            & 45.4 	& 53.6 & 60.7	& 57.7 \\ \hline
Two branches	        & 52.6 & 60.9 & 67.5	& 64.6 \\ 
Three branches      & \textcolor{red}{\textbf{53.9}} & \textcolor{red}{\textbf{62.3}}  & \textcolor{red}{\textbf{69.2}}	& \textcolor{red}{\textbf{66.2}}\\
Four branches	        & 52.4  & 60.7  & 67.6	& 64.6 \\ 
    \hline%\toprule[1pt]
    \end{tabular}
    \vspace{-0.2cm}
      \caption{Study about how many branches are suitable for DEE.}  \label{tab:tab6}
      \vspace{-0.3cm}
\end{table}

\begin{small}
\begin{table}[t]\footnotesize
  \centering
  \renewcommand{\arraystretch}{0.8}
  \begin{tabular}{l|cc|cc}\hline%\toprule[1pt]
    \multirow{2}{*}{Methods} & \multicolumn{2}{c|}{LLCM}& \multicolumn{2}{c}{SYSU-MM01} \\    \cline{2-5}
                         & R-1 & mAP & R-1 & mAP  \\
   \hline

%baseline	& 45.4 	& 53.6 & 60.7	& 57.7  \\ \hline
NL 	& 50.1	& 57.4	& 63.8	& 60.7  \\ 
MFA	&\textcolor{red}{\textbf{51.2}} & \textcolor{red}{\textbf{59.6}} & \textcolor{red}{\textbf{64.7}} & \textcolor{red}{\textbf{62.0}}\\ 
\hline
NL + DEE	& 54.2	& 62.4	& 73.4	& 70.3  \\ 
MFA+ DEE	& \textcolor{red}{\textbf{54.9}} & \textcolor{red}{\textbf{62.9}}  & \textcolor{red}{\textbf{74.7}} & \textcolor{red}{\textbf{71.8}}\\ 
\hline
    %\toprule[1pt]
    \end{tabular}  \vspace{-0.3cm}
  \caption{Comparison with the Non-Local (NL) block.}  \label{tab:tab7}
  \vspace{-0.4cm}
\end{table}
\end{small}

\begin{figure}[t]
\centerline{\includegraphics[height=8.1cm,width=8.2cm]{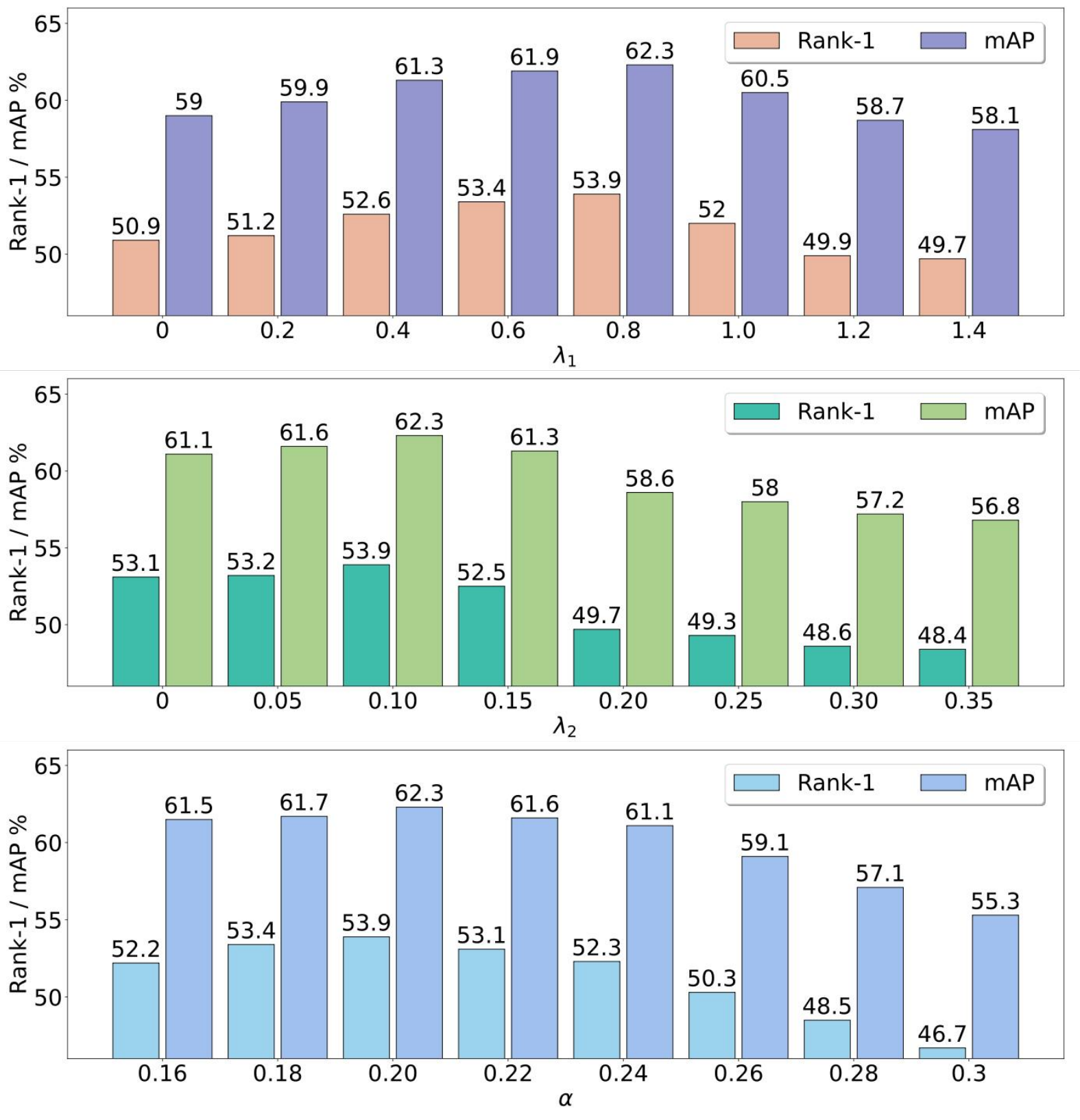}}
\vspace{-0.4cm}
\caption{Influence of different $\lambda_1$, $\lambda_2$ and $\alpha$ values on our LLCM.}  \label{img:img6}
\vspace{-0.5cm}
\end{figure}

%\begin{figure}[t]
%\includegraphics[height=4cm,width=8.5cm]{img7.pdf}
%\caption{Validation results of CQC Loss on the RegDB dataset.}
%\label{img:img4}
%\end{figure}

\begin{figure*}[t]
\centering
\includegraphics[height=6.4cm,width=17.5cm]{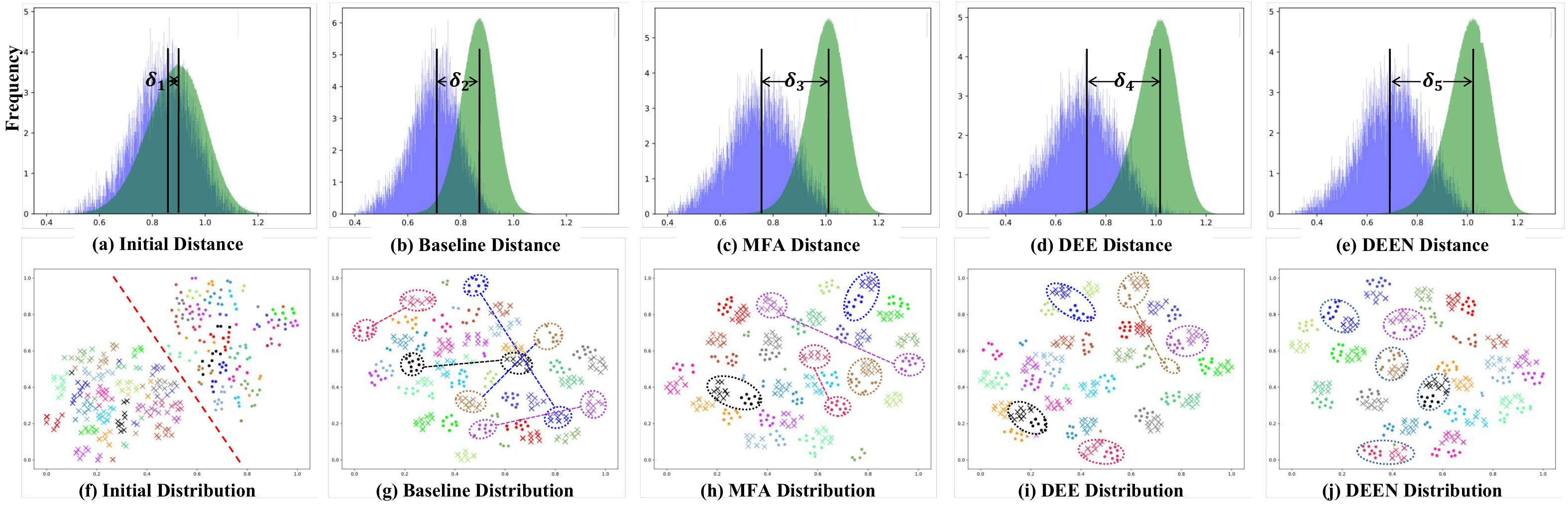}
\vspace{-0.5cm}
\caption{(a-e) show the intra-class and inter-class distances of cross-modality features. The intra-class and inter-class distances are indicated in blue and green colors, respectively. (f-j) show the distribution of feature embeddings in the 2D feature space, where circles and triangles in different colors denote visible and infrared modalities. A total of 20 persons are selected from the test set. The samples with the same color are from the same person. The “dot” and “cross” markers denote the images from the VIS and IR modalities, respectively.}\vspace{-0.3cm}
\label{img:img7}
\end{figure*}

%\begin{figure}[t]
%\includegraphics[height=6cm,width=7.5cm]{img7.pdf}
%\caption{The Rank-10 retrieval results obtained by the baseline and our method on the SYSU-M001 %dataset.}
%\label{img:img7}
%\end{figure}

%\begin{figure}[t]
%\includegraphics[height=8cm,width=8.2cm]{img7.pdf}
%\vspace{-0.3cm}
%\caption{Comparison of different $\lambda_2$ in Eq. (\ref{eq1}) on the RegDB (top row) and SYSU-MM01 dataset (bottom row).}
%%\vspace{-0.5cm}
%\%label{img:img7}
%\end{figure}

%The comparison of DEEN in different layers of backbone.

\textbf{The influence of which stage of ResNet-50 to plug the DEE module.} The proposed DEE can be plugged after any stage of the backbone network. In our experiments, we use ResNet-50 as the backbone, which has five stages: stage-0 to stage-4. We plug DEE after different stages of the ResNet-50 to study how it will affect the performance of the DEEN. As shown in Tab. \ref{tab:tab5},  %It indicates that after stage-3, the proposed DEE is more suitable for generating more diverse embeddings. Based on the above analysis, we plug the DEE into stage-3 of the backbone in if not specified.
 when DEE is plugged after stage-0 to stage-3, the performance gradually increases, which shows the modality gaps become smaller and the generative ability of DEE becomes stronger at deeper layers of the network. When DEE is plugged after stage-3, it can achieve the best results on both LLCM and SYSU-MM01. However, when DEE is plugged after stage-4, the performance drops rapidly because the CPM loss works directly on the embeddings, enlarging the distances between the generated embeddings and the original embeddings, which increases the difficulty of model optimization. Based on the above analysis, we plug DEE after stage-3 of the backbone if not specified.

\textbf{Effectiveness on how many branches are more suitable for DEE.} The proposed DEE module utilizes a multi-branch convolutional block to generate diverse embeddings. Here, we study how many branches are suitable for DEE. As shown in Tab. \ref{tab:tab6}, with the increase of the number of DEE’s branches from 2 to 3, more embeddings are generated to reduce the modality gaps, so the performance gradually increases. However, the increase of performance has an upper limit when the number of branches is more than 3, because DEE generates too many redundant features, which leads to the drop of performance. As a result, DEE with three branches can achieve the best performances both on the LLCM and SYSU-MM01 datasets. It indicates that DEE with 3 branches is more suitable for generating diverse embeddings. Thus, we use 3 branches for DEE if not specified.

\begin{figure}[t]
\includegraphics[height=4.6cm,width=8.5cm]{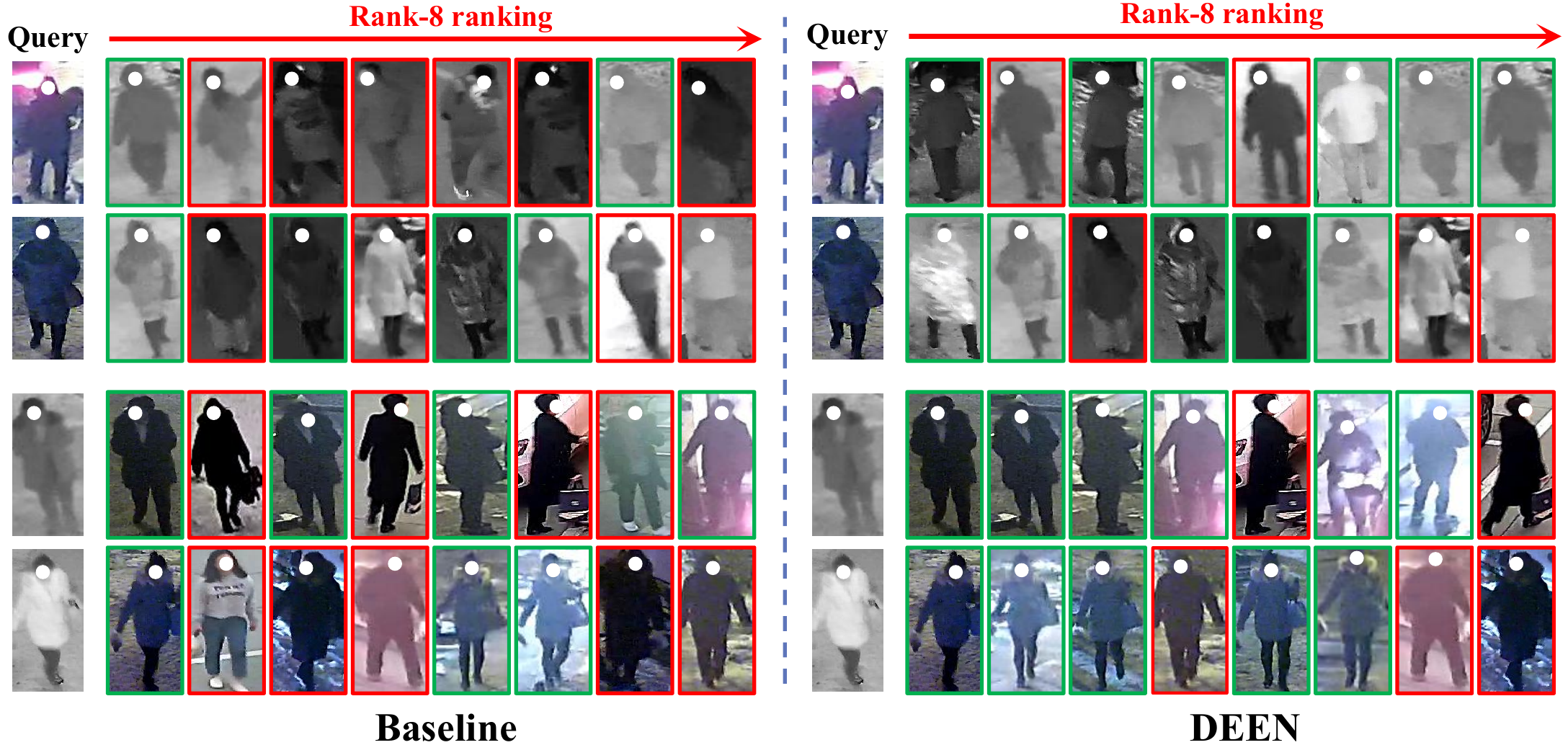}
\vspace{-0.5cm}
\caption{Some Rank-8 retrieval results obtained by the baseline and the proposed DEEN on our LLCM dataset.}
\vspace{-0.4cm}
\label{img:img8}
\end{figure}

\textbf{Comparison with the Non-Local block.} In this paper, we propose a MFA block to mine diverse channel-wise and spatial feature representations inspired by the Non-local (NL) block in \cite{wang2018non}. Thus, we compare these two blocks to investigate which block is more effective. As shown in Tab. \ref{tab:tab7}, the MFA block outperforms the NL block by 1.1\% Rank-1 accuracy and 2.2\% mAP, respectively. The results validate the effectiveness of our MFA block. Moreover, the results also show that the MFA block and the DEE module are complementary for generating diverse embeddings to reduce the modality gaps between the VIS and IR images.

\textbf{The influence of the hyperparameters $\lambda_1$, $\lambda_2$ and $\alpha$.} To evaluate the influence of the three hyperparameters, we give quantitative comparisons and report the results in Fig. \ref{img:img6}. 
As we can see, the best performance is achieved when $\lambda_1$ is set to 0.8, $\lambda_2$ is set to 0.1 and $\alpha$ is set to 0.2, respectively.

\subsection{Visualization}
\textbf{Feature distribution.} To investigate the reason why DEEN is effective, we visualize the inter-class and intra-class distances on our LLCM dataset as shown in Fig. \ref{img:img7} (a-e). Comparing Fig. \ref{img:img7} (c-e) with Fig. \ref{img:img7} (a-b), the means (i.e., the vertical lines) of inter-class and intra-class distances are pushed away by MFA, DEE and DEEN, where $\delta_1$ \textless $\delta_2$ \textless $\delta_3$ and  $\delta_1$ \textless $\delta_2$ \textless $\delta_4$ \textless $\delta_5$. This shows that the intra-class distance of DEEN is significantly reduced compared with the intra-class distance of the initial features (Fig. \ref{img:img7} (a)) and the baseline features (Fig. \ref{img:img7} (b)). Thus, DEEN can effectively reduce the modality discrepancy between the VIS and the IR images. Meanwhile, we also visualize the feature distribution with t-SNE \cite{van2008visualizing} in the 2D feature space in Fig. \ref{img:img7} (f-j), which shows that MFA, DEE and DEEN can effectively discriminate and aggregate feature embeddings of the same person, and reduce the modality discrepancy.

\textbf{Retrieval result.} To further show the effectiveness of DEEN, we also show some retrieval results of DEEN on our LLCM dataset in Fig. \ref{img:img8}. For each retrieval case, the retrieved images with green boxes mean the correct matches corresponding the given query, while the red ones mean the incorrect matches. In general, DEEN can effectively improve the ranking results with more correctly matched images ranked in the top positions than the baseline. 
%To further evaluate the proposed DEEN, we compare the retrieval results obtained by our method with those obtained by the baseline on the LLCM dataset. The obtained Rank-8 ranking results are shown in Fig. \ref{img:img8}. In general, the proposed DEEN can effectively improve the ranking results with more images of green bounding boxes ranked in the top positions. 

\section{CONCLUSION}

In this paper, we propose a novel diverse embedding expansion network (DEEN) in the embedding space for the VIReID task. % which consists of a DEE module and a MFA block. 
The proposed DEEN can generate diverse embeddings and mine diverse channel-wise and spatial embeddings to 
learn the informative feature representations for reducing the modality discrepancy between the VIS and IR images. Moreover, we also provide a challenging low-light cross-modality (LLCM) dataset, which has more new and important features and can further facilitate the research of VIReID towards practical applications. Extensive experiments on the SYSU-MM01, RegDB and LLCM datasets show the superiority of the proposed DEEN over several other state-of-the-art methods.

\section{Acknowledgments}

%This work was supported by the National Key Research and Development Program of China under Grant 2022ZD0160402, and the FuXiaQuan National Independent Innovation Demonstration Zone Collaborative Innovation Platform Project under Grant 3502ZCQXT2022008. %, the National Natural Science Foundation of China under Grants U21A20514, 62176195 and 62071404, and the Open Research Projects of Zhejiang Lab under Grant 2021KG0AB02.

This work was supported by the National Key Research and Development Program of China under Grant 2022ZD0160402, by the National Natural Science Foundation of China under Grant U21A20514, and by the FuXiaQuan National Independent Innovation Demonstration Zone Collaborative Innovation Platform Project under Grant 3502ZCQXT2022008. %, the National Natural Science Foundation of China under Grants U21A20514, 62176195 and 62071404, and the Open Research Projects of Zhejiang Lab under Grant 2021KG0AB02.

%%%%%%%%% REFERENCES
{\small
\bibliographystyle{ieee_fullname}
\bibliography{egbib}
}

\end{document}